# Training, learning and inference: unified dynamics of neural systems

Mian Wang[1,*]

[1] Wuhan Polytechnic University, Wuhan 430023, Hubei, China

[*]Corresponding author: Mian Wang

## Abstract

We define an atomic generation fact, $f=(u,\tau,\omega,z;\rho)$, recording the participating origin, realized transformation, concrete occurrence, generated result, and relation role. Atomic generation facts are compiled into a Generation-Fact Graph (GFG), providing an AI-native, compilable scientific fact substrate in which facts and derived graphs remain connected through their concrete generation histories. On this basis, we establish the GFG-based recursive scientific process: scientific questions guide the analysis of existing GFGs; when analysis, intervention, replay or validation forms a new result, the corresponding execution establishes new generation facts, which enter the next scientific cycle after validation and recompilation.

Using nanoGPT, we establish unified training–learning dynamics. Training is the evolution of a parameter–optimizer system with state and memory: each actual training action enters the current receiving state, and the functional response it produces is jointly conditioned by that state and target-specific update geometry, with finite amplitude and nonlinearity. Learning is the persistent reorganization of distributed functional support by these responses; observable capability formation, maintenance, decline, or recovery arises when target-specific internal states are evaluated across their readout boundaries.

The theory identifies three primary coordinates of training response: the target's current boundary state, target-specific update geometry, and the parameter–Adam receiving state. From these coordinates we derive a second-order target-boundary predictor that operates after the actual update is formed but before the target response is computed, without access to post-update outputs. On fully held-out confirmation runs, it achieved 91.43% target-boundary accuracy and 91.49% macro-averaged recall across remaining correct, declining, remaining incorrect, and recovering.

We further establish that inference is a frozen projection of training–learning dynamics. Component gating and version rollback across thirteen nanoGPT histories showed that frozen inference causally recruits and non-additively combines query-conditioned distributed support formed during training. This relation explains how increased learning formation yields richer projectable support and stronger inference, and derives the organizational conditions realized by Attention: query-conditioned projection and joint combination of distributed learned support. In the studied system, controlled feedback experiments further showed that reinforcement may have double-edged effects.

Finally, experiments in ResNet/CIFAR and diffusion/CIFAR confirmed receiving-state-conditioned training responses, persistent support reorganization, and the frozen, query-conditioned and non-additive projection of learned support during inference, extending the unified dynamics beyond nanoGPT.

## 1. **Introduction**

Computer systems are studied not only through the results they produce, but also through how those results are formed. Over time, several fields have developed mature mechanisms of their own for this purpose: database lineage connects outputs to contributing inputs[1]; Source Maps connect positions in transformed artefacts to positions in their original sources[2]; OpenTelemetry represents distributed operations and their relations[3]; W3C PROV models entities, activities, and relations of use, generation, and derivation[4]; and PyTorch Autograd records executed forward operations and the dependencies required for backpropagation[5,6]. Yet as these mechanisms have matured, each field has encountered questions that demand a more complete account of particular results: why an expected result was absent[7]; how a final artefact passed through multiple transformations[2]; how individual inputs and outputs are related in batched or scatter–gather execution[3]; whether a provenance graph is complete and authoritative at generation time[8,9]; and how training sources, realized computation, recomputation, and individual parameter updates are connected[5,6,10].

Each field has therefore developed a mechanism for describing how results are formed, while the further problems it encounters demand a more complete account of the formation of a particular result. This raises a general question: is there a unified generation relation capable of stating the complete facts of a concrete generation? We define the atomic instance of such a relation as a complete generation fact:

$$f = (u, \tau, \omega, z; \rho) \tag{1}$$

where $u$ is the source information that actually participated, $\tau$ is the transformation actually realized, $\omega$ is the concrete occurrence, $z$ is the outcome formed or explicitly disposed of, and $\rho$ is the relation role.

We prove that the five coordinates are individually irreducible, and that database lineage[1], Source Maps[2], OpenTelemetry[3], W3C PROV[4], PyTorch Autograd[5,6], and canonical $\mathbb{N}[X]$ provenance[11] are exact strict projections of this information-richer structure.

After establishing a method for recording concrete generation, we applied it to real nanoGPT training[12]. We did not begin by selecting loss, gradients, parameters, neurons, attention heads, or any other standard machine-learning object as a predetermined explanation of capability. For each real training run, we first constructed a base training GFG that preserved what actually occurred: which training sources entered which computations, when those computations occurred, and which gradients, optimizer states, and parameter updates they formed. Subsequent capability evaluations, support probes, response measurements, causal interventions, and frozen inferences likewise formed their own generation facts and derived GFGs through real executions, and were precisely connected to preceding GFGs through source-graph identities, original-object identities, and generation paths.

Together, these interconnected GFGs constituted a traceable history of training formation. This made it possible to study training, learning, and inference within the same empirical record, and to investigate the relationships among them. The remainder of this paper presents what we observed, how the resulting mechanisms were identified, and how each conclusion was tested through prospective prediction and causal intervention.

## 2. Model of generation facts

The executable model consists of five components: a capture protocol, execution records, a generation binder, a validated snapshot, and an AI-native Generation-Fact Graph construction layer.

### 2.1 Atomic generation facts

An atomic complete generation fact is

$$f = (u, \tau, \omega, z; \rho) \tag{2}$$

where:

$u$ is the source information that actually participated in the generation described by the fact;

$\tau$ is the transformation realized in that formation;

$\omega$ is the concrete occurrence in which the transformation was realized;

$z$ is the outcome position, instantiated either as an OutcomeSupport or as an ExplicitDisposition;

$\rho$ is the role of $u$ in the relation.

### 2.2 Capture protocol

Let $e$ denote a concrete execution. Before $e$ begins, a capture protocol specifies which runtime records must be captured and how those records provide the coordinates in Eq. (2). The captured records are associated with a declared scope $d$, identified by domain_scope_id. During execution, the generator performs its native computation and produces its ordinary result, while the records required by the protocol are captured as they are produced, either emitted directly by the generator or obtained through synchronous instrumentation.

### 2.3 Establishing generation facts

The source records, transformation references, occurrence records, outcome supports or dispositions, and relation roles delivered under the capture protocol do not constitute generation facts merely by existing separately. A generation fact is established when the records carrying its five elements are jointly bound by a GenerationBinding.

In the executable model, the generation binder follows the fixed capture protocol and binds together $u$, $\omega$, $z$ and $\rho$ from the current execution. Here, $\tau$ is determined by the realized transformation recorded in the concrete occurrence $\omega$, thereby establishing the complete generation fact.

The binding also establishes the correspondence between the generation fact and its outcome position. If $z$ is an OutcomeSupport, the fact describes the formation of that result or a declared part of it. If $z$ is an ExplicitDisposition, the fact records that no outcome support was formed within the declared scope.

The complete generation state established for execution $e$ under scope $d$ is

$$\Gamma_d(e) = \text{multiset}\{f_i\} \tag{3}$$

The state is a multiset because repeated generation facts remain independently significant and must not be collapsed by set semantics.

### 2.4 Multi-stage generation

When the result of one stage participates in a later stage, the Core represents it as a GeneratedOrigin. A binding in the later stage may use that generated origin in the $u$ position while retaining its connection to the earlier stage.

Thus, multi-stage generation is formed by connecting atomic facts through generated origins:

$$f_i \xrightarrow{\text{GeneratedOrigin}} f_j \tag{4}$$

**2.5 Validation and delivery**

The records and GenerationBindings produced by an execution are stored in a Snapshot. Let $S_d(e)$ denote the executable state submitted for validation. Its validation condition is

$$V_d\left(S_d(e)\right) = \text{valid} \tag{5}$$

Validation checks that every GenerationBinding, within the same declared scope, validly binds the records and role information that determine $u$, $\tau$ ,$\omega$ ,$z$ and $\rho$, and that all records required by the capture protocol are covered by bindings. For each binding, the Core also verifies its evidence, the generator's authorization and the successful generation operation that jointly delivered the binding and its evidence. The Snapshot additionally fixes the identities of the protocol, implementation and execution environment.

A Snapshot satisfying these conditions is delivered as a ValidatedSnapshot.

**2.6 Construction of the Generation-Fact Graph**

A ValidatedSnapshot is used to construct an AI-native Generation-Fact Graph,

$$G_e = (V_F, V_O, E_I, E_R; \Sigma) \tag{6}$$

where $V_F$ contains the established atomic generation facts while preserving their identities and multiplicities;

$V_O$ contains the concrete occurrences admitted by the occurrence catalog, including occurrences that realize no atomic fact;

$E_I$ records the exact realizes_fact incidence between concrete occurrences and atomic generation facts;

$E_R$ preserves validated typed relations with their native endpoint types and identities;

$\Sigma$ fixes the schema and registries under which the graph is constructed.

The GFG thereby retains complete atomic generation facts while organizing them with concrete occurrences and typed relations into a structure directly traversable by AI. The constructed graph is validated against its source records and frozen graph contracts before delivery.

## 3. The GFG-based recursive scientific process

To investigate complex formation mechanisms using accumulated generation facts, we established the GFG-based recursive scientific process shown in Fig. 1.

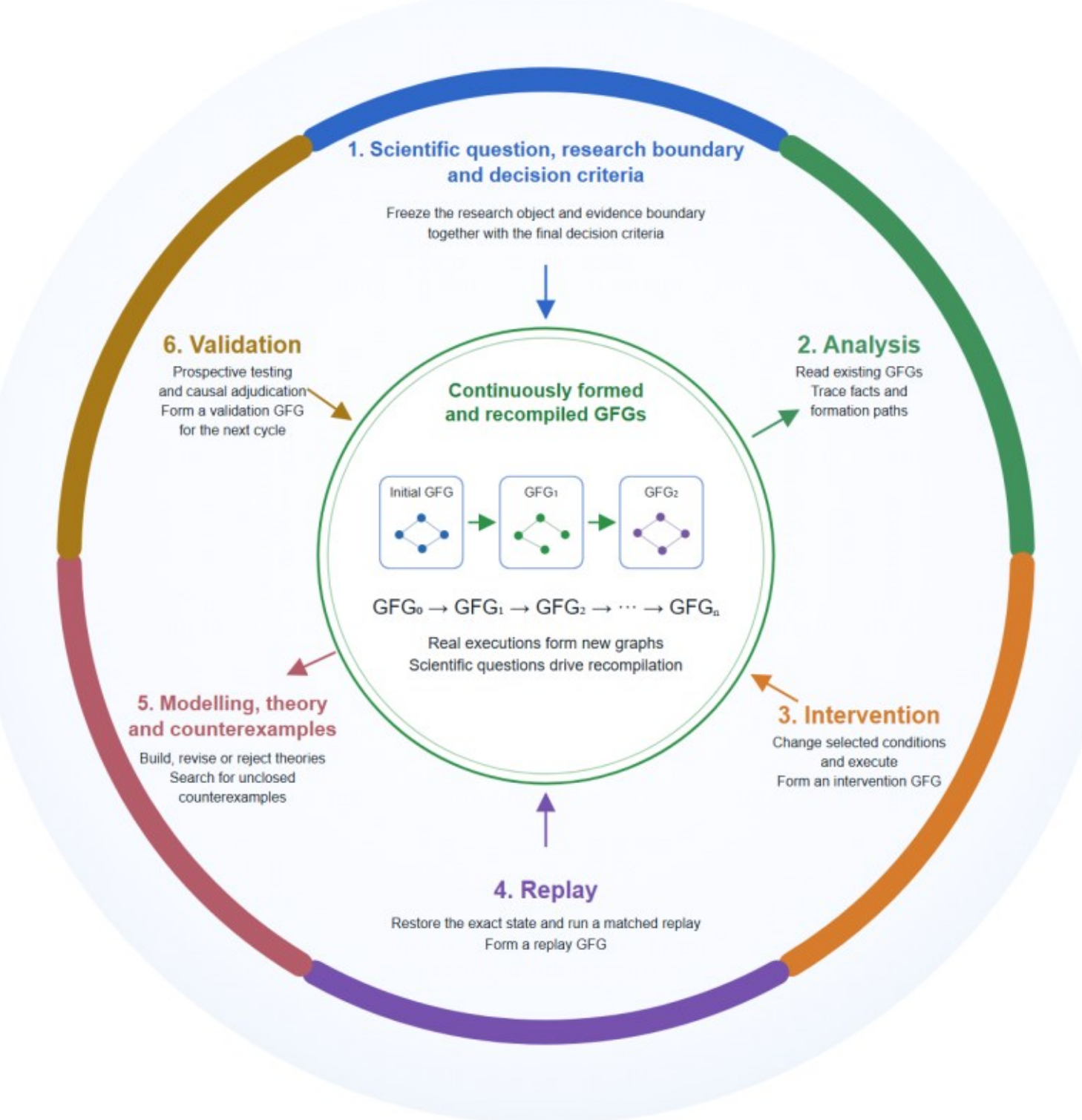

**Figure 1 | The GFG-based recursive scientific process**

Scientific investigation proceeds recursively over an accumulating family of GFGs. A scientific question determines how existing GFGs are queried and recompiled; analysis traces established facts and formation paths, while interventions and replays produce new executions and therefore new generation facts. Modelling, counterexample construction and validation operate on this accumulated evidence, and their realized results can in turn be compiled into subsequent GFGs. The resulting process is recursive: $GFG_0 \rightarrow GFG_1 \rightarrow GFG_2 \rightarrow \cdots$, with each scientific cycle both reading from and contributing to the evolving scientific fact state.

The experiments that follow were conducted largely within this scientific process.

## 4. Discovering the dynamics of training and learning

We used thirteen actual nanoGPT training histories to test, in sequence, the relations constituting the training–learning mechanism. Each experiment was grounded in the complete facts provided by the corresponding training GFG and obtained its machine results through actual execution, intervention or replay. We recorded the operation performed, the machine-observed result, the proposition supported or falsified, and the contribution of that result to the final theory.

### 4.1 Experiments

#### TL-E01 — Cross-run state sufficiency and post-formation trajectories

We first asked whether familiar scalar observables were sufficient to describe capability dynamics across runs. We compared states with similar loss, accuracy, absolute training step, individual target margins or final-layer normalization gain and examined their subsequent trajectories. None of these quantities was sufficient: states that appeared similar under each description could lead to different capability futures. Capability also did not become permanently stable at its first appearance. After formation, it could be maintained, decline and later recover.

This experiment falsified the view that capability is determined by a single scalar progress variable or by absolute training time. It established that training must instead be described as a stateful process whose relevant state retains information not contained in conventional learning curves.

#### TL-E02 — Optimizer pauses and clipping negative controls

We then intervened on the evolution of the parameter–optimizer system. Pausing parameter or optimizer evolution delayed capability formation by 1800, 800 and 1100 steps in three independently sealed predictions. By contrast, changing the gradient-clipping threshold did not reproduce the same delays.

These interventions distinguished the effective evolution of the parameter–optimizer system from a simpler explanation based only on gradient magnitude or clipping. They showed that capability formation depends on the realized training action entering and changing a system with persistent parameter and optimizer memory.

**TL-E03 — Full/skip branches, receiving-state exchange and reciprocal response**

To test whether a training update has a fixed effect independent of where it is applied, we constructed full/skip matched causal branches from identical pre-update states and batches, and separately conducted receiving-state-exchange experiments in which corresponding updates were applied to different receiving states. The matched branches included executing the complete update, skipping it, and separating parameter and optimizer-state contributions.

The same update produced different functional responses when applied to different receiving states. In local notation,

$$J_A \Delta\theta \neq J_B \Delta\theta \tag{7}$$

Conversely, states that appeared similar under the existing macroscopic description could respond differently to an otherwise matched update. The effect of a training action is therefore not an intrinsic property of the update alone. It is jointly determined by the update and the state that receives it.

**TL-E04 — Multi-amplitude paths of realized updates**

The local-response experiments left one question unresolved: can a first- or second-order local approximation be extrapolated to a complete training update?

Starting from the same pre-update state, we applied the same realized update at a frozen set of amplitudes:

$$\alpha \in \{0, 0.125, 0.25, 0.5, 0.75, 1\} \quad (8)$$

The complete response curves exhibited near-linearity and four recurrent nonlinear morphologies—saturation, acceleration, turnback and sign reversal (Fig. 2).

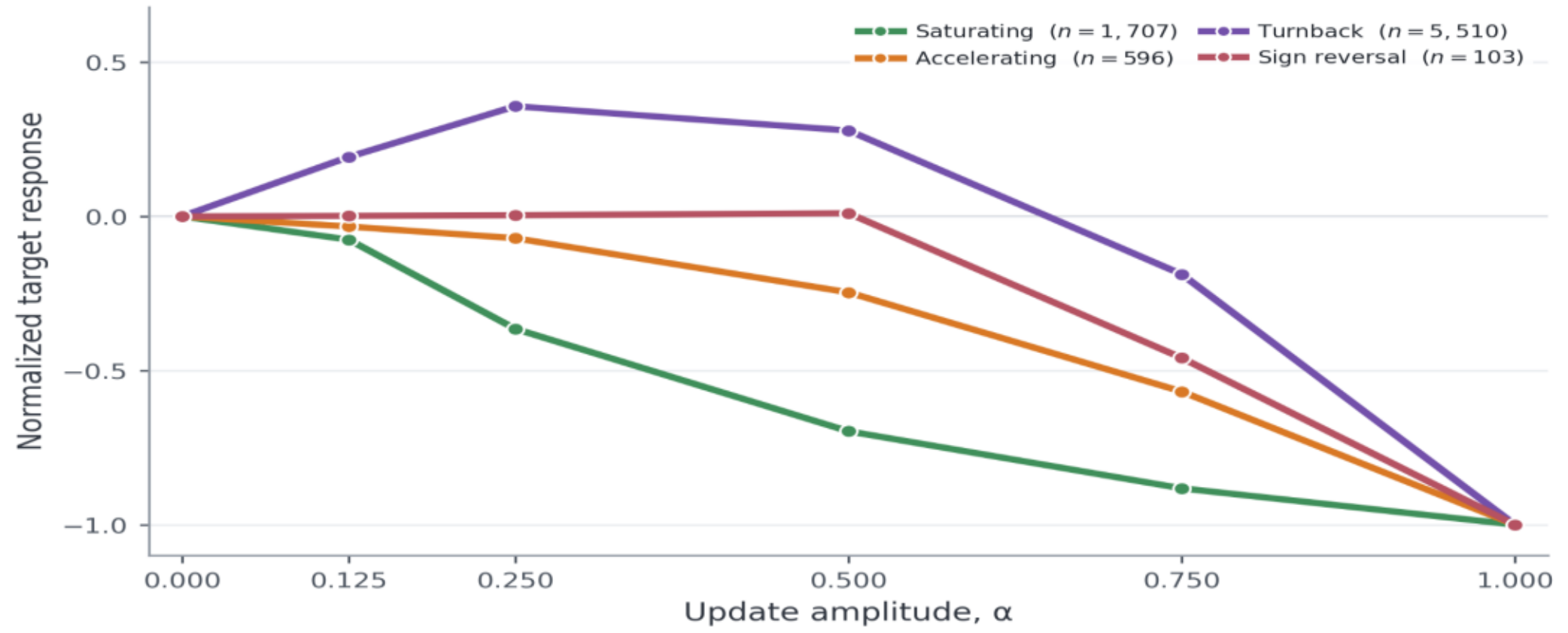

**Figure 2 | Finite-amplitude nonlinear response morphologies.**

Fixed local *J* or *J/K* approximations described some small-amplitude responses but did not reliably recover the endpoints of complete updates.

This experiment falsified the proposition that the functional responses to complete training actions can generally be recovered by fixed linear or quadratic local extrapolation, and established finite-amplitude nonlinear functional response as an object of study.

**TL-E05 — Analysis of response-conditioning factors**

Having established that the responses were nonlinear, we tested whether this nonlinearity varied without structure or was conditioned by states already present before the update occurred. The experiment compared seven candidate information groups using run-isolated matching, ablation and cross-run evaluation.

Three stable primary conditioning factors were identified:

- **the target's current boundary state;**
- **target-specific update geometry;**
- **the parameter–Adam receiving state.**

Once these three information classes were available, natural-history variables and past response curves provided no stable independent gain. This does not imply that history has no effect; rather, its effective influence was already represented primarily in the current parameter–Adam receiving state.

Training responses are therefore neither governed by a single globally fixed function nor composed of unrelated random curves at individual steps. They are nonlinear responses jointly conditioned by the current receiving state and target-specific update geometry.

**TL-E06 — CSRG single- and pairwise-component gating**

We next examined the internal structure supporting an acquired capability. At each materialized training state, CSRG-4C-v1 performed two baseline evaluations, four single-component gating interventions and six pairwise-component gating interventions. For every specific target, it compared the logits, margins and predictions produced under these interventions.

Different targets and training states exhibited different patterns of component necessity, pairwise substitution, backup, synergy and failure tolerance. The same capability could not be consistently attributed to one fixed component. When one component was disabled, other components could sometimes take over its function, but not always.

This experiment established distributed functional support: a capability is jointly sustained by multiple identity-preserving components whose division of responsibility, substitutability and redundancy vary with the training state.

**TL-E07 — Support reallocation across realized updates**

We then compared the unupdated state, $\alpha=0$, with the fully realized update state, $\alpha=1$, from the same receiving state, thereby directly measuring how an actual training action changed the support network.

Across 72 realized-update sections, the experiment established 1656 target-group support transitions and 1636 valid support-allocation transitions. The primary supporting component changed in 228 cases. Total update magnitude had a Spearman correlation of 0.881 with the magnitude of support reallocation, while support-reallocation magnitude correlated at 0.765 with the absolute change in capability. Mean reallocation magnitude was 0.2080 when capability changed, compared with 0.0340 when it remained unchanged.

However, the update magnitude of an individual component did not reliably determine whether that component subsequently gained or lost functional responsibility. The direction of support change was jointly determined by the current support configuration and the complete update.

Learning therefore cannot be defined merely as a change in parameters. Learning is the persistent reorganization of distributed functional support produced by training responses.

**TL-E08 — Identity-aligned target margins and boundary crossing**

Finally, we examined how internal responses and support reorganization give rise to observable capability outcomes. For every specific target, the experiment aligned its identity, correct class, competing class, logits and margin before and after the update, and then determined whether that target crossed the readout boundary between correct and incorrect behaviour.

A training update could alter the parameters, Adam state, functional response and support structure while leaving the externally observed result unchanged if the specific target did not cross its corresponding boundary. Only when an identity-aligned target state crossed that boundary did one of four observable outcomes arise:

- remaining correct;
- changing from correct to incorrect;
- remaining incorrect;
- changing from incorrect to correct.

This experiment established the final link between internal formation and external capability outcomes. Capability is not a direct synonym for arbitrary internal change; it is the result produced when the internal functional state of a specific target is evaluated through its corresponding readout boundary.

### 4.2 The training–learning theory established by the experiments

Together, the eight experiments establish the following mechanistic chain:

Actual training action

↓

Conditioning by the current parameter–optimizer receiving state

and target-specific update geometry

↓

Target-specific finite-amplitude nonlinear functional response

↓

Persistent reorganization of distributed functional support

↓

Target-specific readout boundary

↓

Capability formation, maintenance, decline or recovery

We therefore formulate the training–learning theory:

Training is the evolution of a parameter–optimizer system with state and memory. Each actual training action enters the current receiving state; the functional response it produces is jointly conditioned by that state and target-specific update geometry, and is finite-amplitude and nonlinear. Learning is the persistent reorganization of distributed functional support by these responses. Only when the internal state of a specific target crosses its corresponding readout boundary does this process become observable as capability formation, maintenance, decline or recovery.

## 5. Direct prediction of target-level learning outcomes

If the mechanism identified in Section 4 is valid, then once an actual parameter update has been formed but before the target response occurs, it should be possible to predict, from the target's current boundary state, target-specific update geometry, and current receiving state, whether the update will cause the target to remain correct, decline, remain incorrect, or recover.

**5.1 Second-order target-boundary prediction**

For evaluation target $e$, designated target token $y$, and competing token $c$, define the pre-update margin

$$g_{(e,c)}(\theta_t) = \ell_y(\theta_t, x_e) - \ell_c(\theta_t, x_e) \tag{9}$$

This margin is $F1$: the target's current position relative to its readout boundary. The actual parameter update $\Delta\theta_t$, acting on this particular target, provides $F3$. Its effect is evaluated at the current receiving state $\theta_t$, which provides the parameter component of $F5$. At this event boundary, the Adam state has already participated in forming $\Delta\theta_t$.

The predicted post-update margin is

$$\hat{g}_{e,c,t+1} = \underbrace{g_{e,c}(\theta_t)}_{F1} + \underbrace{Dg_{e,c}(\theta_t)[\Delta\theta_t] + \frac{1}{2}D^2 g_{e,c}(\theta_t)[\Delta\theta_t, \Delta\theta_t]}_{F3 \text{ acting in } F5}. \tag{10}$$

The algorithm therefore implements

$$\text{predicted post-update boundary} \approx F1 + \text{the response produced when } F3 \text{ enters } F5. \tag{11}$$

The designated target token was predicted to rank first after the update when

$$\min_{c \neq y} \hat{g}_{e,c,t+1} > 0. \tag{12}$$

Combining the observed pre-update boundary with the predicted post-update boundary produced four target transitions: remained correct, correct to incorrect, remained incorrect, and incorrect to correct.

**5.2 Predictive results**

On the frozen confirmation split, the predictor correctly determined the post-update target-boundary result in 4652 of 5088 cases. Target-boundary accuracy was 91.43%, balanced accuracy was 92.17%, and macro-averaged recall across the four transitions was 91.49%. The recall for each transition is summarized in Table 1.

**Table 1 | Recall of target-boundary transitions**

| Target-boundary transition | Recall |
| --- | --- |
| **Remained correct** | 87.18% |
| **Correct → incorrect** | 98.91% |
| **Remained incorrect** | 95.90% |
| **Incorrect → correct** | 83.95% |

Across all 12 eligible, fully materialized training runs, the same formula correctly predicted 14069 of 15264 target-update outcomes, giving 92.17% accuracy and 91.30% macro-averaged recall.

The algorithm was derived directly from the three primary coordinates identified in Section 4 and achieved a macro-averaged recall of 91.49% across the four capability transitions. The predictive result thereby validates the training–learning theory in the executed nanoGPT system.

## 6. From learning formation to inference

The preceding experiments established how actual training actions reorganize distributed functional support and become observable capabilities. One relation remained unresolved: after training has formed a capability, how does that formation participate in a new inference occurrence? We therefore tested whether frozen inference causally recruits and combines the functional support formed during training.

### 6.1 Inference is a frozen projection of training–learning dynamics

We tested thirteen nanoGPT training histories at 52 checkpoints spanning capability formation, decline and recovery. Four comparisons examined whether frozen inference used the functional support formed during training. The results are summarized in Table 2.

**Table 2 | Evidence for frozen-inference projection**

| Test | Result | Relation established |
|---|---|---|
| **Exact training versions and component gating** | Exact trained versions were read; gating changed the full logits. | Inference uses components formed by training. |
| **Target-wise component-gating responses** | At every formed checkpoint, 23 target groups produced 23 distinct response patterns. | Different targets recruit different distributed supports. |
| **Single- and pair-component gating** | All 138 component-pair-by-target comparisons at every checkpoint were non-additive. | Inference jointly combines multiple supports. |
| **Pre-formation rollback and restoration** | 52/52 rollbacks reduced accuracy; restoration exactly recovered the original logits. | Inference causally depends on the component versions formed during capability formation. |

Every parameter read during inference matched a parameter-version object recorded in the corresponding training GFG. The registered components produced nonzero outputs, and gating them changed the full logits. The inference result therefore depended on concrete components and parameter versions formed in the recorded training history.

Different targets used these components differently. For each target, the pattern of logit changes caused by gating the four components indicated how its function was distributed among them. In all thirteen runs, this pattern changed from pre-formation to formation. At every formed checkpoint, the 23 target groups produced 23 different patterns. Frozen inference therefore did not use one fixed component allocation for every target; the current query recruited a target-specific distribution of support.

We next asked whether the recruited components contributed independently. If they did, the effect of gating two components together would equal the sum of their separate effects. At each checkpoint, six component pairs were tested across 23 target groups, producing 138 comparisons. Every comparison departed from independent addition. The inference result therefore depended on the joint participation of multiple supports.

The rollback intervention established that these supports were produced by learning. At each formed checkpoint, one component was replaced by the version of the same component immediately preceding capability formation, while the query and the rest of the formed network were kept unchanged. All 52 rollbacks changed the full logits and the responses of all 23 target groups; accuracy decreased in all 52 cases. Restoring the formed component version exactly restored both the parameter hashes and the original logits. The inference result therefore depended causally on the specific component versions formed during training.

Together, the four results establish the following relation:

$$\text{training-formed component versions} \xrightarrow{\text{current-query recruitment}} \text{target-specific distributed support} \xrightarrow{\text{joint combination}} \text{inference result} \quad (13)$$

The current query recruits and combines functional support already formed by training, while the learned state remains persistently unchanged. Inference is therefore a frozen projection of training–learning dynamics.

### 6.2 The scale effect follows from the expansion of projectable functional support

This relation also explains why increased training scale can produce stronger inference.[13] Inference can project only functional organization that learning has already formed. As training forms and reorganizes more usable functional support, the learned state can produce richer query-conditioned projections and stronger combinations during inference:

increased training scale

↓

increased learning formation

↓

richer projectable functional support

↓

stronger inference

Training scale therefore affects inference through learning formation. The rollback result demonstrates this dependence directly: with the architecture and query unchanged, replacing a formed component by its earlier version weakened the inference result, while restoring the formed version restored the result exactly.

### 6.3 Why Attention succeeds

Attention realizes the organizational conditions required by this projection:

$$\mathrm{Attention}(Q,K,V)=\mathrm{softmax}\left(\frac{QK^{\top}}{\sqrt{d}}\right)V. \tag{14}$$

The query–key relation constructs the active projection for the current occurrence, while the values carry the distributed contents recruited by that projection.[14] Attention allows the same learned state to form different active support configurations for different queries and to combine them without persistently modifying the learned parameters. Its success follows from this correspondence: Attention makes the distributed functional support formed during training dynamically projectable by the current query.

### 6.4 Reinforcement as a potentially double-edged feedback process

A foundational achievement of reinforcement learning is that it uses the environmental consequences of action for subsequent learning. A system produces an action through inference, receives the environmental feedback formed by that action and updates its policy accordingly, thereby changing its future behaviour.[15]

Within the present theory, reinforcement learning closes the loop between inference and subsequent training: inference projects already formed functional support into action, and the consequences of that action can return as new training actions. When positive consequences repeatedly favour particular projected outputs, these training actions can preferentially reorganize functional support toward the processes that produced them, forming a positive-feedback path. In controlled experiments, increasing the concentration of correct positive feedback produced a strict dose-ordered increase in support for the reinforced capability and a corresponding dose-ordered decline in the margins of unreinforced capabilities, with sufficiently concentrated feedback eventually producing observable capability loss. Rebalancing subsequent feedback produced substantial recovery across all twelve seeds while preserving the reinforced capability.

**Table 3 | Experimental evidence for the double-edged reinforcement process**

| Evidence | Formal result | Implication |
|---|---|---|
| Dose ordering | Target support ↑ and other-skill margins ↓ strictly in 12/12 seeds (mean Spearman $\rho = +1.000 / -1.000$). | Feedback concentration systematically redirects functional support. |
| Endpoint trade-off | Exclusive vs balanced: target support share +9.67 pp; other-skill accuracy −39.06 pp. | Support amplification can accompany collateral capability loss. |
| Rebalancing recovery | Other-skill accuracy recovered to 99.48%; target remained 100% in 12/12 seeds. | Redistributing feedback largely reverses the trade-off. |
| Fresh reproduction | All 12 formal seeds were freshly re-executed; the primary results were reproduced. | The result is reproducible beyond the main-run summary. |

These findings do not argue against reinforcement learning. Rather, they suggest that sustained asymmetric feedback may benefit from monitoring functional-support concentration and the margins of unreinforced capabilities, together with adjustment of feedback concentration, duration and balance when collateral weakening emerges.

## 7. Cross-system validation

To test whether the identified relations were specific to nanoGPT, we repeated the frozen training–learning and inference protocols in two structurally different systems: a ResNet classifier and a diffusion model, both using CIFAR data. Across three formal seeds per system, TL-G01 and TL-G02 recovered receiving-state-conditioned functional responses and persistent support reorganization, including 504 independently checked diffusion-response records. INF-G01 further recovered exact state preservation, query-conditioned support recruitment, non-additive combination and dependence on training-formed component versions in all six formal seeds. Complete protocols, GFGs, checkpoints and independent validators are provided in the public evidence archive. These results exclude a nanoGPT-specific explanation and support the cross-system scope of the proposed dynamics.

## 8. Conclusion

This work ultimately reveals training, learning and inference not as separate processes, but as successive relations within a unified dynamics of neural systems. Training is the evolution of a parameter–optimizer system with state and memory: each actual training action enters the current receiving state, and the functional response it produces is jointly conditioned by that state and target-specific update geometry, with finite amplitude and nonlinearity. Learning is the persistent reorganization of distributed functional support caused by these responses; when target-specific internal states are evaluated against their respective readout boundaries, this reorganization becomes observable as capability formation, maintenance, decline or recovery. Inference does not continue this persistent reorganization. Instead, the current query projects and jointly combines functional support already formed by learning, while the learned state remains persistently unchanged. Experiments in ResNet/CIFAR and diffusion/CIFAR further confirmed these relations, extending the unified dynamics beyond nanoGPT.

The GFG-based recursive scientific process is not another stage in this loop, nor is it limited to machine learning. It operates on an accumulating family of GFGs that provide a general, AI-native and compilable scientific fact substrate for concrete generation. Real executions establish generation facts that can be compiled, according to different scientific questions, into derived GFGs while preserving identities and generation paths; subsequent analyses, interventions, replays and computations establish further generation facts and support continued recompilation. The closed

training–learning–inference dynamics established here is therefore one scientific discovery produced through this recursive scientific process, rather than the boundary of what the process can be used to study.

edn (MIT Press, 2018).

## Methods

### Generation-fact evidence

Before each execution, a frozen capture protocol specified how runtime records corresponded to the five coordinates of an atomic generation fact and to any typed relations required by the experiment. Captured records were bound into atomic facts, validated within their declared scope and compiled into a GFG. Subsequent queries, replays, interventions and analyses operated on this validated evidence. When an analysis produced a new result, it established a new generation fact linked to its input evidence through GeneratedOrigin; merely selecting an existing subgraph produced no new fact. The complete experimental contracts, implementations and machine-verifiable evidence are provided in the accompanying repository.

### Structural projection experiments

We tested whether five established mechanisms—database lineage, Source Maps, OpenTelemetry, W3C PROV and PyTorch Autograd—could be recovered from complete generation facts. For each mechanism, the task, native output, projection rule and comparison criteria were frozen before execution. The native mechanism and the projection from the validated complete generation state were produced independently and compared in identity, field content, order and multiplicity. Strictness was tested using paired generation states that differed while producing the same native view.

Classical $\mathbb{N}[X]$ provenance was tested separately on 13 relational-algebra tasks. One implementation computed provenance during relational execution, whereas another derived it from the recorded generation facts. Standardized polynomials were compared by variables, coefficients and exponents, followed by projection to bag multiplicity, Boolean existence and positive Boolean expressions.

**Training–learning experiments**

We analysed thirteen independently executed nanoGPT training histories. Their GFGs preserved training inputs, forward and backward computations, gradients, Adam states, parameter updates, parameter versions and capability evaluations. All comparisons retained object identity and execution order; future capability facts were excluded whenever a pre-update or historical state was being evaluated.

**Cross-run state sufficiency (TL-E01)**

At multiple historical cut points, states from different runs were matched using loss, accuracy, absolute training step, target margin and final-layer normalization gain, separately and in combination. Their withheld subsequent capability trajectories were then compared. States with similar summaries but divergent futures were retained as counterexamples. The same trajectories were audited after initial capability formation to identify subsequent maintenance, decline and recovery.

**Optimizer interventions (TL-E02)**

Exact checkpoint states were replayed under interventions that paused effective parameter–optimizer evolution and under negative controls that changed gradient-clipping thresholds while preserving the remaining training procedure. Formation time and the subsequent capability trajectory were measured relative to the corresponding uninterrupted execution.

**Receiving-state dependence (TL-E03)**

For matched states from different runs, the native training step was branched into skip and full-update executions. Realized parameter updates and Adam innovations were then applied reciprocally to the other receiving state while parameter identities, continuation data and random-number opportunities were aligned. Functional response was measured as the difference between each update branch and its corresponding skip branch. This design tested whether the effect of an actual training action depended only on the update itself or also on the state receiving it.

**Finite-amplitude response (TL-E04)**

The same realized parameter update was applied to an identical pre-update state at the frozen amplitudes specified in Eq. (15).

$$\alpha \in \{0, 0.125, 0.25, 0.5, 0.75, 1\}. \tag{15}$$

At each amplitude, target margins, predictions and functional-support measurements were recomputed. Complete response paths were compared with first-order and second-order local approximations derived around the unmodified state.

**Response-conditioning factors (TL-E05)**

Each target-specific response was represented by its margin curve over the frozen amplitude grid. Seven groups of pre-response information were evaluated: current boundary state, competitor structure, target-specific update geometry, functional-support state, parameter–Adam receiving state, natural history and previously observed response history. Features were standardized using development runs only. Semantically identical targets were matched across different runs, and factor contributions were assessed through incremental inclusion, block deletion and alternative inclusion orders. Runs, rather than individual targets, formed the outer separation and statistical unit.

**Capability-support redundancy (TL-E06)**

At each materialized checkpoint, CSRG-4C-v1 executed two ungated baselines, four single-component gates and six paired-component gates over the four registered residual components. Gating set the selected component output to zero before residual addition and did not modify the stored model state. Target logits and margins from these executions were used to calculate component necessity, paired backup, effective support, support concentration and single- and double-component failure slack. All probes at a checkpoint used the same restored parameter version and evaluation inputs.

**Support reallocation (TL-E07)**

For every eligible response section, we compared the support state obtained without applying the recorded update, $\alpha = 0$, with that obtained after applying the complete update, $\alpha = 1$, to the same receiving state. For component allocation vectors $a_0$ and $a_1$, reallocation magnitude was defined as

$$R = \frac{1}{2} \| a_1 - a_0 \| \tag{16}$$

The audit also recorded changes in necessity, backup, effective support, failure slack and primary-support identity.

**Target-boundary crossings (TL-E08)**

Evaluation targets were aligned across the pre-update and post-update executions by their preserved identities rather than by array position or numerical similarity. For each target, the margin was defined as the correct-target logit minus the largest competing logit. Its sign before and after the update determined one of four transitions: remaining correct, correct to incorrect, remaining incorrect or incorrect to correct. This target-level ledger connected internal functional change and support reallocation to observable capability outcomes.

**Actual-update boundary prediction (TL-P01)**

We tested the predictive sufficiency of the three response coordinates using twelve nanoGPT training runs. Each prediction was made after the actual parameter update $\Delta\theta_t$ had been formed, but before the corresponding post-update target logits, margins or predictions were computed.

For the correct target $y$ and each competing target $c$, the post-update target boundary was predicted by

$$\hat{g}_{y,c} = g_{y,c}(\theta_t, x) + Dg_{y,c}(\theta_t, x)[\Delta\theta_t] + \frac{1}{2}D^2 g_{y,c}(\theta_t, x)[\Delta\theta_t, \Delta\theta_t] \tag{17}$$

The pre-update boundary represented the target's current boundary state; the directional response represented the target-specific geometry of the actual update; and $\theta_t$, together with the already formed $\Delta\theta_t$, represented the receiving parameter–Adam state. A target was predicted to be correct only when $\hat{g}_{y,c} > 0$ for every competitor. The four transitions were derived mechanically from pre-update and predicted post-update correctness; no separate classifier was trained. All 15264 target instances were evaluated under the same rule, with complete runs separated between development and confirmation sets.

**Frozen-inference projection (INF-E01)**

We tested inference using exact parameter checkpoints and CSRG evidence from all thirteen training histories. For each run, four checkpoints were selected mechanically from the previously recorded capability series: immediately before formation, at formation, at the largest post-formation decline and after recovery. Checkpoint selection was completed before any new inference or rollback result was read.

At each checkpoint, the frozen model performed real CUDA forwards on held-out inputs without persistent parameter or optimizer updates. Two ordinary forwards were required to produce byte-identical logits. During each forward, the inputs, outputs, parameter-version identities and call order of four registered residual components were captured. Four single-component gates and six paired-component gates were then executed to determine target- and query-conditioned component effects and non-additive interactions.

At the formation checkpoint, each component was separately replaced by its exact pre-formation parameter version while all other components remained unchanged. The hybrid model performed a new inference, after which the formation-version bytes were restored and exact output recovery was verified. The experiment tested whether frozen inference used exact training-formed parameter versions, causally recruited distributed support, changed its support effects with the query and combined component contributions non-additively.

**ResNet/CIFAR cross-system validation (TL-G01)**

Three independently seeded ResNet-18 models were trained on CIFAR-100 using

SGD with momentum. At sealed checkpoints, realized updates were replayed across frozen amplitudes and exchanged between parameter and optimizer receiving states; all sixteen coalitions of four residual stages were executed before and after each update to measure target responses and functional-support reorganization.

**Diffusion/CIFAR cross-system validation (TL-G02)**

Three independently seeded DDPM-style U-Nets were trained on CIFAR-10 using AdamW. Targets were fixed by image, diffusion timestep and noise-occurrence identity; realized-update replays, receiving-state exchanges and all sixteen coalitions of four U-Net support routes were used to measure target responses and persistent support reorganization.

**Cross-system frozen inference projection (INF-G01)**

Exact trained checkpoints from TL-G01 and TL-G02 were evaluated without further training. Query-conditioned component gates, pair interactions and rollback to exact pre-learning component versions tested support recruitment, non-additive combination, dependence on training-formed versions and preservation of the learned state during inference.

**Selective positive feedback (RL-E05)**

After four capabilities had been formed, twelve matched seeds were divided into selective-feedback, balanced-feedback and frozen branches with identical initial parameter–AdamW states and matched budgets. Functional support was measured through all sixteen component coalitions, and all fifteen non-empty component-version rollback subsets were tested to adjudicate support concentration, capability trade-offs and their temporal relation.

**Feedback dose and recovery (RL-E06)**

From the same mastered receiving state, matched branches received balanced, mild, high or exclusive positive feedback for 3,200 updates, with a frozen branch as a no-update control. Exact states at update 800 initiated rebalanced and repair branches, which were compared with continued exclusive feedback using capability outcomes, continuous margins and functional-support trajectories.

**Validation, statistics and reproducibility**

Experimental protocols, admissible inputs, interventions, evaluation rules and stopping conditions were frozen before the corresponding target results were read. Deterministic projection experiments were adjudicated by exact equality rather than statistical approximation. Training analyses preserved complete runs as the independent experimental units; individual targets, checkpoints and GFG nodes were not treated as independent biological or experimental replicates. Where cross-run matching was used, uncertainty was estimated by clustering over runs or unordered run pairs.

Prediction performance was evaluated on complete held-out runs using accuracy, balanced accuracy, per-transition recall, macro-averaged recall and complete confusion matrices. Frozen-inference experiments retained all thirteen runs, and reinforcement-learning experiments retained all twelve paired seeds. No failed run, target or seed was removed after outcome inspection.

All source objects, parameter versions, updates, interventions and derived outcomes were identified by content hashes and connected to their generating occurrences. Experimental outputs were sealed before hidden or post-update results were accessed. Independent checkers recomputed the reported measurements from the sealed records, verified source and output hashes, and rejected missing identities, invalid event order, future leakage or disagreement between the recorded facts and reconstructed results. Derived analyses established new generation facts linked to their exact source evidence through GeneratedOrigin. Exact software versions, frozen contracts, execution manifests, verification programs and experiment-specific reproduction instructions are provided in the accompanying repository.

**Experimental instruments**

Four experimental instruments were developed for the training–learning experiments. CSRG-4C measured how individual components and component pairs supported each target. Realized-update causal forks separated the effect of an actual training update from that of the state receiving it. Finite-amplitude update paths measured the complete functional response produced by applying the same update at different amplitudes. The identity-aligned target-boundary ledger connected each target's internal change to whether it remained correct, declined, remained incorrect or recovered.

## The foundational role of the GFG-based recursive scientific process

Although the unified training–learning–inference dynamics may itself be a foundational scientific discovery, the more general methodological contribution may be the GFG-based recursive scientific process that produced it. To test this foundational role directly, we used the same process to address long-horizon temporal credit assignment through four successive experiments.

**Consequence binding and temporal credit (RL-E01)**

With the receiving parameter–optimizer state, actions and physical consequences held identical, changing only consequence binding or temporal credit changed the actual parameter update and post-update logits in all 48 matched causal forks. Correct binding and credit re-formed a reversed policy in all twelve formal seeds, whereas disrupting either relation prevented the same formation.

**GFG-guided temporal-credit discovery (RL-E02)**

Without being supplied the true credit relation, GFG formation-path retrieval reduced a 64-action history to nine candidates while retaining all six functional actions and three formation ancestors without causal effect. Matched causal replay assigned zero credit to these passengers and exactly recovered credit signs and pairwise interactions. Training with the discovered credit achieved the same held-out result as the hidden oracle, including 96.48% terminal success.

**Recursive optimization of credit discovery (RL-E03)**

Early actions were propagated through a genuine multistage formation chain, with every retained action passing through at least 36 state transformations before the terminal consequence. The credit-discovery computation itself was then captured and compiled into a validated GFG. Recompiling its formation structure preserved exact credit within the frozen $10^{-12}$ tolerance while reducing native replay transitions by 90.64% and producing a 2.38-fold end-to-end speedup, including construction of the credit-discovery GFG.

### Stochastic long-chain temporal credit (RL-E04)

Occurrence-addressed stochastic inputs were introduced into every transition of the long formation chain. Matched replay separated action-contingent credit from realized environmental variation, recovered conditional credit and its sign, and preserved zero credit for formation ancestors without causal effect. Training with the discovered credit achieved 94.82% terminal success and 98.62% functional-action accuracy on held-out policies.

## AI-assisted scientific execution

General-purpose AI systems, including OpenAI Codex and ChatGPT, were used as scientific agents within the GFG-based recursive scientific process. The first author defined the scientific questions, research boundaries and decision criteria, introduced the principal theoretical hypotheses and conceptual transitions, and retained authority over all scientific decisions. Under the first author's direction, the AI systems used their existing domain knowledge and the accumulated GFG evidence to perform much of the subsequent analysis; implement, initiate and monitor experimental designs; conduct interventions and matched replays; search for counterexamples; and execute validation procedures. All reported machine results were produced by the recorded native programs and execution environments rather than generated by the AI systems. No conclusion was accepted unless it satisfied the predefined criteria and was supported by native execution, generation-fact evidence and independent verification. The first author reviewed the resulting evidence and accepts responsibility for the accuracy, integrity and conclusions of the work.

## Data availability

The generation-fact evidence, frozen experimental records, validation results and supporting data underlying this study are available in the accompanying public evidence archive at https://doi.org/10.5281/zenodo.22032772. The archive is linked to the frozen experimental repository and identifies the evidence associated with each experiment reported in the manuscript.

## Code availability

The frozen protocols, generation-fact and GFG implementations, executable experiments, independent validators and reproduction entry points are available at https://github.com/wind342/gfg-training-learning-inference-experiments, tag paper-experiments-cross-system-feedback-release, commit 36dab5ce347dbbdac157ef23205f556606d18294.

## Funding

This research received no specific grant from any funding agency in the public, commercial or not-for-profit sectors.

## Author contributions

Mian Wang conceived the study; developed the atomic generation-fact model, the Generation-Fact Graph, the GFG-based recursive scientific process, the experimental instruments, and the unified theory of training, learning and inference; defined the scientific questions, research boundaries and decision criteria; introduced the principal theoretical hypotheses and conceptual transitions; directed the research programme; analysed and interpreted the validated results; developed the prospective predictions and theoretical deductions; prepared the figures; and wrote the manuscript.

## Competing interests

The author declares no competing interests.

## Additional information

Correspondence and requests for materials should be addressed to Mian Wang (wangmian945@gmail.com).